\documentclass{egpubl}
\usepackage{egsgp2023}
 
\SpecialIssuePaper         

\CGFStandardLicense

\usepackage[T1]{fontenc}
\usepackage{dfadobe}  

\usepackage{cite}  
\usepackage{algorithmic}
\usepackage[ruled,vlined]{algorithm2e}
\SetKwInput{KwInput}{Input}
\SetKwInput{KwOutput}{Output}
\usepackage{array}
\usepackage{adjustbox}
\usepackage{graphicx}
\usepackage{amsmath}
\usepackage{amssymb}
\usepackage{booktabs}
\usepackage{color}
\usepackage{multirow}

\BibtexOrBiblatex
\electronicVersion
\PrintedOrElectronic

\renewcommand{\paragraph}[1]{\noindent \textbf{#1}}

\usepackage{egweblnk} 

\title{3D Keypoint Estimation Using Implicit Representation Learning}

\author[Zhu et al.]
{\parbox{\textwidth}{\centering 
        Xiangyu Zhu$^{1\dagger}$\orcid{0000-0001-7222-4677},
        Dong Du$^{1,3}$\thanks{Equal contribution.}\orcid{0000-0001-5481-389X},
        Haibin Huang$^{4}$\orcid{0000-0002-7787-6428},
        Chongyang Ma$^{4}$\orcid{0000-0002-8243-9513},
        Xiaoguang Han$^{1,2}$\thanks{Corresponding author: hanxiaoguang@cuhk.edu.cn}\orcid{0000-0003-0162-3296}
    }
    \\
    {\parbox{\textwidth}{\centering 
        \textsuperscript{1}SSE, CUHKSZ \quad
        \textsuperscript{2}FNii, CUHKSZ \quad
        \textsuperscript{3}City University of Hong Kong \quad
        \textsuperscript{4}Kuaishou Technology
        \\
        }
    }
}

\begin{document}


\newcommand{\point}{\mathbf{p}}
\newcommand{\normal}{\mathbf{n}}

\maketitle
\begin{abstract}

In this paper, we tackle the challenging problem of 3D keypoint estimation of general objects using a novel implicit representation. Previous works have demonstrated promising results for keypoint prediction through direct coordinate regression or heatmap-based inference. However, these methods are commonly studied for specific subjects, such as human bodies and faces, which possess fixed keypoint structures. They also suffer in several practical scenarios where explicit or complete geometry is not given, including images and partial point clouds. Inspired by the recent success of advanced implicit representation in reconstruction tasks, we explore the idea of using an implicit field to represent keypoints. Specifically, our key idea is employing spheres to represent 3D keypoints, thereby enabling the learnability of the corresponding signed distance field. Explicit keypoints can be extracted subsequently by our algorithm based on the Hough transform. Quantitative and qualitative evaluations also show the superiority of our representation in terms of prediction accuracy.



\end{abstract}

\section{Introduction}
\label{sec:introduction}

In this paper, we study the challenging and under-explored problem of 3D keypoint estimation for general shapes. As a key component in many downstream tasks, an accurate and robust 3D keypoint estimation method can provide useful clues for various applications, including 3D object detection \cite{mian2006three,liu2020smoke,barabanau2019monocular}, object tracking \cite{schall20083d,choi2010real,bugaev2018combining}, shape matching \cite{zou2008surface,bugaev2018combining,wang2018learning}, and shape registration \cite{barnea2008keypoint,li2015towards,bueno2016detection}. 

\begin{figure}[!t]
	\centering
	\includegraphics[width=0.83\linewidth]{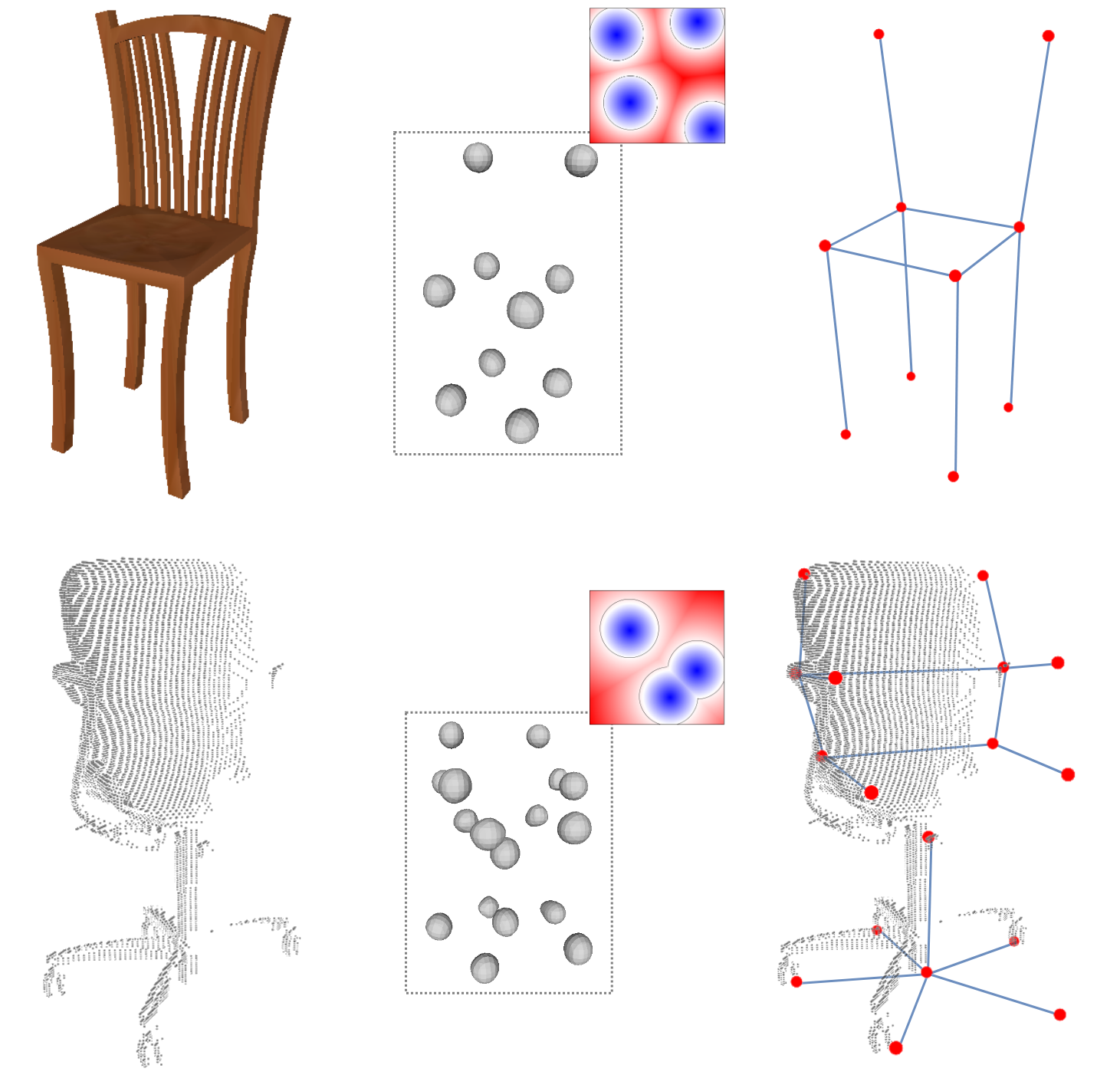}
	\caption{3D keypoint estimation via our implicit sphere learning. Given a single image or partial point cloud, we learn the SDF of keypoint spheres, and the sphere meshes are extracted for the final keypoint estimation. To enhance visualization, we draw lines connecting the keypoints.}
	\label{fig:teaser}
        \vspace{-7mm}
\end{figure}

Although existing methods have demonstrated great success in the detection of facial landmarks as well as human body joints \cite{berretti20113d,creusot2011automatic,papandreou2017towards,papandreou2017towards,geng2021bottom}, they are designed for shapes with consistent structures. It is commonly not easy to extend such methods for 3D keypoint estimation of general shapes which usually present diverse geometric topologies and irregular numbers of keypoints. 
Recently, You et al. \cite{you2020keypointnet} proposed the first large-scale 3D keypoint dataset of 16 general object categories in ShapeNet \cite{chang2015shapenet} and established a benchmark for the task of keypoint prediction. All the methods evaluated in \cite{you2020keypointnet} focus on complete point cloud input, where the keypoint prediction can be converted into a classification task for each point. However, explicit or complete geometry is typically expensive to obtain. For example, the input for keypoint estimation can be images or partial point clouds. This makes classification-based methods infeasible since the expected 3D keypoints cannot be explicitly obtained from the input.

To tackle the challenge of keypoint estimation for general objects, alternative methods leverage deep neural networks and can be roughly grouped into two categories, i.e., point coordinate regression \cite{fan2017point} and heatmap inference \cite{oberweger2018making}. Methods that directly regress spatial coordinates of keypoints are straightforward but increase the risk of overfitting. Moreover, both order and number of keypoints generally need to be fixed for the design and implementation of the network, which is unreasonable for shapes with varying structures and topologies. On the other hand, the heatmap representation is often proposed for 2D keypoint estimation.
The consumption of calculation and storage increases significantly for 3D scenarios, leading to low-resolution heatmap prediction and poor accuracy of keypoint estimation. Neither point regression nor heatmap inference is adequate to generate accurate 3D keypoints in irregular and unordered cases. 

Inspired by the recent success of accurate 3D reconstruction with implicit shape learning \cite{mescheder2019occupancy,park2019deepsdf,chen2019learning,chibane2020implicit}, we propose a novel implicit representation for 3D keypoints estimation. Specifically, 3D keypoints are represented as the centers of distinct spheres with a user-specified radius, then the signed distance field (SDF) of these spherical shapes can be inferred using classical implicit learning methods. Given a point cloud or a single-view image, we adopt a deep neural network to learn the SDF field and extract explicit sphere meshes, followed by keypoint extraction using the 
Hough transform. With this new formulation, we are able to not only handle uncertain number and order properties of general object keypoints but also to improve the performance of keypoint estimation for incomplete point cloud or image inputs, as shown in Figure~\ref{fig:teaser}. Furthermore, we explore the semantic keypoint prediction with implicit learning using the proposed stacked unsigned distance field (UDF), in order to benefit applications that require semantic information.

We conduct experiments about 3D keypoint regression on the KeypointNet dataset \cite{you2020keypointnet} to compare our method with two alternative representations, i.e., the point coordinate and heatmap. Comparisons for complete point cloud, partial point cloud, and single-view image input settings are presented respectively. Both quantitative and qualitative results demonstrate the superiority of our implicit representation for 3D keypoint estimation.  

Our contributions can be summarized as follows:
\begin{itemize}
	\item We introduce the continuous implicit field as a sparse point representation for the first time and propose a consistent 3D keypoint estimation framework for general objects with various topologies and geometry.
	\item With our implicit representation, we also propose a novel architecture that can generate keypoints with semantic labels. 
	\item We conduct extensive experiments on 3D keypoint estimation with various inputs including complete point clouds, partial point clouds, and single-view images, which demonstrate the superiority of our implicit representation.
\end{itemize}

In the following sections of this paper, we will first review related work about 3D keypoint detection and estimation, as well as implicit representation learning in Section~\ref{sec:related_work}. Then, we propose our implicit keypoint representation and architecture for keypoint learning, extraction, and semantic prediction in Section~\ref{sec:method}. Next, we evaluate and compare our method with existing approaches both quantitatively and qualitatively in Section~\ref{sec:experiments}. Lastly, we summarize our method and discuss its limitations and future work in Section~\ref{sec:conclusion}.

\section{Related Work}
\label{sec:related_work}

\paragraph{Keypoint detection.}
3D keypoint saliency detection, which picks up keypoints from a full point cloud, has been a classical task for many downstream applications, such as object detection, pose estimation, shape matching, and registration. Traditional methods mainly utilize hand-crafted geometric features to select the most salient keypoints, but they either ignore the semantic information of keypoints or tend to generate misaligned keypoints \cite{novatnack2007scale,zhong2009intrinsic,sun2009concise,tombari2010unique,sipiran2011harris,khoury2017learning}. Li et al.~\cite{li2019usip} pioneer a learning-based 3D keypoint detector, named USIP. However, USIP takes advantage of probabilistic Chamfer loss which may greatly enhance the repeatability of inferred keypoints. Whereafter, Wei et al.~\cite{wei2021multi} attempt to jointly learn the 3D keypoint saliency and correspondence to improve accuracy. Recently, Fernandez et al.~\cite{fernandez2020unsupervised} propose an unsupervised method to learn aligned 3D keypoints by decomposing keypoint coordinates into low-rank non-rigid shape registration. This approach is suitable for the detection of similar shapes but cannot perform well on general objects with various topologies and geometry. Shi et al.~\cite{shi2021skeleton} improve the unsupervised detector with the guidance of skeletons and a proposed composite Chamfer distance. In contrast to these detection-based methods, our method focuses on keypoint generation of general objects,  where the inputs can be incomplete (e.g., partial point clouds and single-view images).

\paragraph{Keypoint estimation.}
Although 3D keypoint detection has achieved great success, it is not suitable for acquiring full keypoints from incomplete inputs. Most of the 3D keypoint regression methods are designed for specific object categories with consistent topologies, such as human faces \cite{eskimez2019noise} and human bodies \cite{kocabas2019self,cheng2019occlusion,doersch2019sim2real,dabral2019multi,yuan2021simpoe}. The keypoint generation of general objects remains challenging since there are diverse topologies and geometric structures in general objects. Recently, He et al. \cite{he2020pvn3d} introduce a voting network for 3D keypoint estimation for point cloud input. Zhou et al. \cite{zhou2018unsupervised} propose an unsupervised domain adaptation method for 3D keypoint prediction from a single depth scan or image. Suwajanakorn et al.~\cite{suwajanakorn2018discovery} also explore an end-to-end geometric reasoning method for the discovery of latent 3D keypoints without supervision. However, this unsupervised method takes pose estimation as a downstream target, and it may not generate useful keypoints for other applications like shape deformation. Vasconcelos et al.~\cite{vasconcelos2019structured} also utilize the domain knowledge for keypoint estimation of general objects, but the performance suffers from the limited dataset. Afterward, You et al.~\cite{you2020keypointnet} provide the first large-scale dataset of annotated keypoints for 16 general object categories. In this paper, we utilize this dataset and propose a unified architecture for 3D keypoint estimation of general objects.

\begin{figure*}[t]
	\centering
	\includegraphics[width=0.98\linewidth]{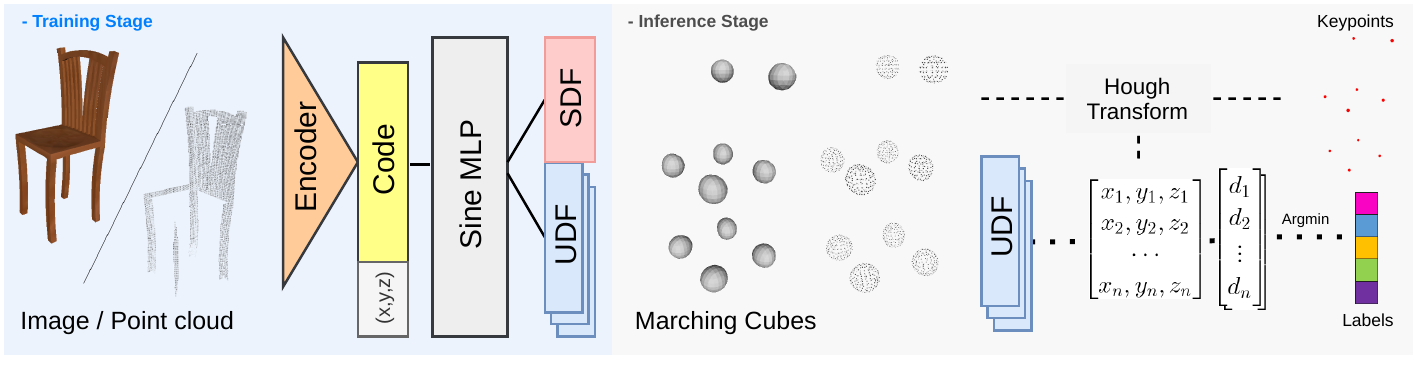}
	\caption{Overview of our implicit keypoint estimation framework. In the inference stage, we extract keypoints from the learned SDF and fetch their coordinates to generate the corresponding semantic labels by the learned stacked UDF.}
	\label{fig:pipeline}
 \vspace{-5mm}
\end{figure*}

\paragraph{Implicit representation learning.}
There are various representations for 3D shape learning, such as volumes~\cite{choy20163d}, point clouds~\cite{fan2017point}, and implicit fields~\cite{mescheder2019occupancy,park2019deepsdf,chibane2020implicit}. However, it is not effective or proper to represent 3D keypoint as volumes. Besides, directly regressing 3D point coordinates is not reasonable since it requires a fixed number of points. Inspired by 2D heatmap regression~\cite{pfister2015flowing,bulat2016human,oberweger2018making}, an alternative keypoint representation is heatmap~\cite{pavlakos2017coarse}, but the consumption of calculation and storage increases extremely for 3D scenarios, leading to low-resolution 3D heatmap prediction and poor accuracy of keypoint estimation. As implicit learning has exhibited its great power for 3D reconstruction, we adopt it for 3D keypoint learning in this paper and validate its superiority in our experiments.

\section{Method}
\label{sec:method}

In this section, we formally introduce our method, including an implicit representation of 3D keypoints and a framework to extract 3D keypoints from various forms of inputs based on the proposed representation. Specifically, we formulate 3D keypoints as implicit spheres represented by SDF and train a deep neural network to infer SDF from various inputs,  such as complete/partial point clouds and single-view images (Section~\ref{sec:3.1}).  We then utilize Marching Cubes~\cite{lorensen1987marching} to obtain explicit spheres' surfaces and estimate centers of spheres with Hough transform algorithm~\cite{hough_sphere} as our final keypoints (Section~\ref{sec:3.2}). Furthermore, we introduce stacked UDF learning for semantic keypoint prediction (Section~\ref{sec:3.3}).

\subsection{Implicit Keypoint Estimation}
\label{sec:3.1}

\subsubsection{Implicit Keypoint Network}

Implicit representation such as SDF has demonstrated its advantages in shape reconstruction with various topologies and geometric structures~\cite{mescheder2019occupancy,park2019deepsdf,chen2019learning,chibane2020implicit}. In this work, we introduce SDF for 3D keypoint representation to handle irregular and unordered keypoints of general objects. Specifically, we expand a keypoint $\mathbf{p}_i$ to a \emph{keypoint sphere} $\partial B(\mathbf{p}_i, r)$ with a user-specified radius $r$ ($r$ is empirically set as 0.08 and is fixed in all of our experiments) where $B(\mathbf{p}_i, r) = \{x \in \mathbb{R}^3|  \left \|x - \mathbf{p}_i \right \|_2 \leq r \}$, 
and define \emph{keypoint spheres} $\mathcal{S}$ for K keypoints $\{\mathbf{p}_i\}_{i=1}^K$ as:
\begin{align*}
    \mathcal{S} = \partial \left ( \bigcup_{i=1}^K B(\mathbf{p}_i, r) \right )
\end{align*}
We adopt the SDF of the keypoint spheres to encode keypoints position for our proposed network, with the definition of SDF as:
\begin{align}
    f(\point): \mathbb{R}^3 \rightarrow s,
\end{align}
where $\point\in\mathbb{R}^3$ is an arbitrary point in the space, and $s=\text{sign}(\point)\cdot d$. Here, $d$ represents the distance from $\point$ to the closest point of the sphere surface. We set $\text{sign}(\point)$ 1 for the points outside the spheres and -1 for the inside. According to recent work \cite{sitzmann2020implicit}, $f$ should also satisfy the following Eikonal equation:
\begin{align} \label{equ:grad}
    \left \| \nabla f(\point) \right \| &= 1, \quad \forall \point \in \mathbb{R}^3, \textit{a.e.} \\
    f(\point) &= 0, \quad \forall \point \in \mathcal{S},
\end{align}
where $\mathcal{S}$ is the keypoint spheres. Our key insight is that we can encode keypoints implicitly with an SDF function $f$. The function is appropriate for an arbitrary number of keypoints and can be well approximated by a neural network function $f_{\theta}$.

We now bridge the 3D keypoint estimation and various types of inputs of 3D general objects (e.g., images and point clouds) by modeling $f_{\theta}$ conditioned on input from the specified space $\mathcal{X}$. Given an observation $x\in\mathcal{X}$, the function takes $(\point, x)\in\mathbb{R}^3\times\mathcal{X}$ to output an SDF value $s$, which can be formulated as:
\begin{align}
    f_\theta: \mathbb{R}^3 \times \mathcal{X} \rightarrow s.
\end{align}
We regard this function as our implicit keypoint network and utilize advanced neural architectures to optimize the parameters $\theta$.


\subsubsection{Network Training}

Our network adopts the encoder-decoder architecture used in DeepSDF~\cite{park2019deepsdf} to learn the implicit field defined w.r.t keypoint spheres. Given the observation of a point cloud or a single-view image, we randomly sample points in the 3D space (i.e., $\left[-1, 1\right]^3$) and fetch them into our network to obtain SDF values. As shown in Figure~\ref{fig:pipeline}, we utilize different encoders for different inputs (PointNet~\cite{qi2017pointnet} for point clouds and ResNet~\cite{he2016deep} for single-view images), and use a multi-layer perceptron (MLP) as the decoder of SDF. Positional encoding is also applied for each sampled point $\point$ before concatenating it with the observation features, to help the network learn high-frequency components of the input position. Following the work of NeRF~\cite{mildenhall2020nerf}, the positional encoding function we use is:
\begin{align}
    \psi(\point) = (\sin(2^0\pi \point), \cos(2^0\pi \point), ..., \sin(2^N\pi \point), \cos(2^N\pi \point)).
\end{align}
The function $\psi$ is applied separately to each coordinate value of $\point$. In our experiments, we set $N=6$.

As mentioned in SIREN \cite{sitzmann2020implicit}, SDF learning benefits from high-frequency features. Therefore, we adopt a sinusoidal activation function, i.e., $\sigma(\cdot) = \sin(\omega \ast \cdot)$, where $\omega=30$ is a specified constant in \cite{sitzmann2020implicit}. We also follow SIREN to initialize the weights of our decoder MLP. 


In our experiment, we adopt SDF loss, gradient loss, and normal loss to supervise the training. Specifically, given an observation $x$ and data pairs $\left\{(\point_i, s_i)\right\}$ of the queried points and the corresponding SDF values, the SDF loss is defined as the $L_1$ difference between the predicted values and the ground truth:
\begin{align}
    L_{\textit{SDF}} = \sum_{x,i} |f_\theta(\point_i, x) - s_i|.
\end{align}
According to Eq.~(\ref{equ:grad}), the norm of $\nabla_{\point_i} f_\theta({\point_i}, x)$ should be restricted to $1$ over $\mathbb{R}^3$. For a point on the sphere surface $\mathcal{S}$, $\nabla_{\point_i} f_\theta(\point_i, x)$ equals to the normal vector $\normal_i$ at $\point_i$, if $\normal_i$ can be defined here. Thus we add two losses associated with the gradient of our network:
\begin{align}
    L_{\textit{grad}} &= \sum_{x,i} | \left \| \nabla_{\point_i} f_\theta(\point_i, x) \right \| - 1|, \\
    L_{\textit{normal}} &= \sum_{x,\point_i \in \mathcal{S}} (1 - \frac{ \nabla_{\point_i} f_\theta(\point_i, x) \cdot \normal_i }{ \left \| \nabla_{\point_i} f_\theta(\point_i, x) \right \|  }).
\end{align}
$L_{\textit{normal}}$ is used to penalize the cosine similarity between $\nabla_{\point_i} f_\theta(\point_i, x)$ and $\normal_i$.

Our final objective of the training is a weighted sum of the three terms:
\begin{align}
    L = \lambda_1 L_{\textit{SDF}} + \lambda_2 L_{\textit{grad}} + \lambda_3 L_{\textit{normal}},
\end{align}
where $\lambda_1$, $\lambda_2$, $\lambda_3$ are the corresponding weights. More details are given in the experimental setup (Section~\ref{sec:experimental_setup}). Note that the positional encoding, sinusoidal activation, and losses associated with the gradient are combined to use for improving the smoothness of keypoint spheres, which can improve the accuracy of the subsequent keypoint estimation from the inferred SDF.

\subsection{Keypoint Extraction}
\label{sec:3.2}

To obtain 3D keypoints from the learned SDF, we utilize the Marching Cubes (MC) algorithm~\cite{lorensen1987marching} to first extract the keypoint spheres mesh from the inferred SDF. However, these spheres might intersect each other, hindering the subsequent keypoint extraction. Notice that sphere detection from a point cloud is a well-studied problem, where Hough transform can be used to acquire spheres from point clouds efficiently and robustly~\cite{hough_sphere}. Inspired by this idea, we take the vertices of intersected spheres as input and utilize a Hough transform-based method to extract the distinct spheres as well as their centers. Note that the keypoint extraction task can be simplified and performed on every mesh-connected component of the output mesh, and the underlying spheres should possess the same radius, which greatly reduces the complexity of the calculation. Therefore, we propose Algorithm~\ref{alg:kp} to extract the keypoints of an unknown number.

In Step 1 of Algorithm~\ref{alg:kp}, we follow the standard Hough transform to voxelize the bounding box of the input point cloud (i.e. one connected component) with a grid size $d$ and perform sphere center voting for all input points. In Step 2, we find out possible clusters containing sphere centers by clustering bins, whose votes are beyond a given threshold $N_{\text{vote}}$. Then we select points with the maximum votes in each cluster as candidate sphere centers. 

In our experiment, the candidate centers calculated in Step 2 are sometimes inaccurate. Therefore, in Step 3, we utilize the nearest points to update the positions of these centers at most $N_{max}$ times. Specifically, we formulate it as a best sphere matching problem (Eq.~(\ref{eq:var})) with the minimum variance for a given point, which has an analytical solution~\cite{best_sphere} given by Eq.~(\ref{eq:center}):

\begin{align}
    \min_{c_L} \sum_{i=1}^{N_L} (\left \|x_i^L - c_L \right \|^2_2 - \frac{1}{N_L} \sum_{j=1}^{N_L} \left \|x_j^L - c_L \right \|^2_2 )^2 \label{eq:var},
\end{align}

\begin{align}
    c_L = \bar{X_L} + \frac{1}{2} \textit{Cov}(X_L)^{-1} \gamma \label{eq:center}.
\end{align}

\begin{align}
    \bar{X_L} = \frac{1}{N_L} \sum_{i=1}^{N_L} x_i^L ,~ \textit{Cov}(X_L) = \frac{1}{N_L} \sum_{i=1}^{N_L} (x_i^L - \bar{X_L}) (x_i^L - \bar{X_L})^T, \nonumber
\end{align}

\begin{align}
     \gamma = \frac{1}{N_L} \sum_{i=1}^{N_L} (x_i^L - \bar{X_L}) (x_i^L - \bar{X_L})^T (x_i^L - \bar{X_L}). \nonumber
\end{align}
where $X_L = \{x_i^L\}_{i=1}^{N_L}$ is the given point cloud and $c_L$ is the optimal center with the minimal variance defined in Eq.~(\ref{eq:var}).

\begin{algorithm}
  \KwInput{$P=\{p_i\}_{i=1}^{N_P}$ }
  \KwOutput{$\{c_i\}_{i=1}^K$}
  \textbf{Step 1.}\\ centers of bins $\{b_i\}_{i=1}^{N_b}$, length $d$ $\gets$ Voxelize bounding box of $P$, \\ set vote($b_i$) = 0, $\forall i$. \\
  \For{$p_i$ in P}{
    vote($b_j$) += 1, $\forall b_j, \left \|b_j - p_i \right \|_2 \leq d/2$
  }
  \textbf{Step 2.} \\ Clusters $\{B_i\}_{i=1}^{N_B} \gets$ Clustering $\{b_j| \text{vote}(b_j) > N_{\text{vote}}\}$ \\
  $c_i^{(0)} \gets \arg\max_{b_j} \{\text{vote}(b_j), b_j \in B_i\}, i=1,2,...,N_B$ \\
  \textbf{Step 3.} \\
  \For{$k=0$ \KwTo $N_{\text{max}}$ }{
    $P_i^{(k)} \gets \left \{p_j \in P \big | \left \|p_j - c_i^{(k)} \right \|_2 \leq \left \|p_j - c_l^{(k)} \right \|_2 \forall l \neq i \right \}, \forall i=1,2,...,N_B$ \\
    $c_i^{(k+1)} \gets $Eq$(\ref{eq:center}) \big |_{X_L = P_i^{(k)}},  \forall i=1,2,...,N_B $ \\
    \If{$\max_{i} \left \|c_i^{(k+1)} - c_i^{(k)} \right \|_2 < \epsilon $}{\textbf{break}}
  }
  \textbf{Step 4.} \\
  \eIf{exists $c_i, c_j$ such that $\left \|c_i - c_j \right \|_2 < \text{radius}$ }
  {Merge all such $c_i, c_j$ with average position, back to \textbf{Step 3.} }
  {return $\{c_i\}_{i=1}^K$}
\caption{Keypoint Extraction}
\label{alg:kp}
\end{algorithm}

It should be noted that in Step 3, there is a possibility of erroneously grouping some close points into different spheres, such as extracting two center points from an ellipsoid-like shape (a non-standard sphere type in our experiments). To rectify this, in Step 4, we merge these close points by calculating their mean position and then return to Step 3 to ensure accurate keypoint extraction.

For the implementation of keypoint estimation, we set the hyper-parameters $d=1/32$, $\text{radius}=0.08$, $\epsilon=0.01$, $N_{\text{vote}}=80$, $N_{\text{max}}=10$. These hyper-parameters remain constant throughout all experiments, which achieves stable performance. 

\begin{figure}[t]
	\centering
	\includegraphics[width=1\linewidth]{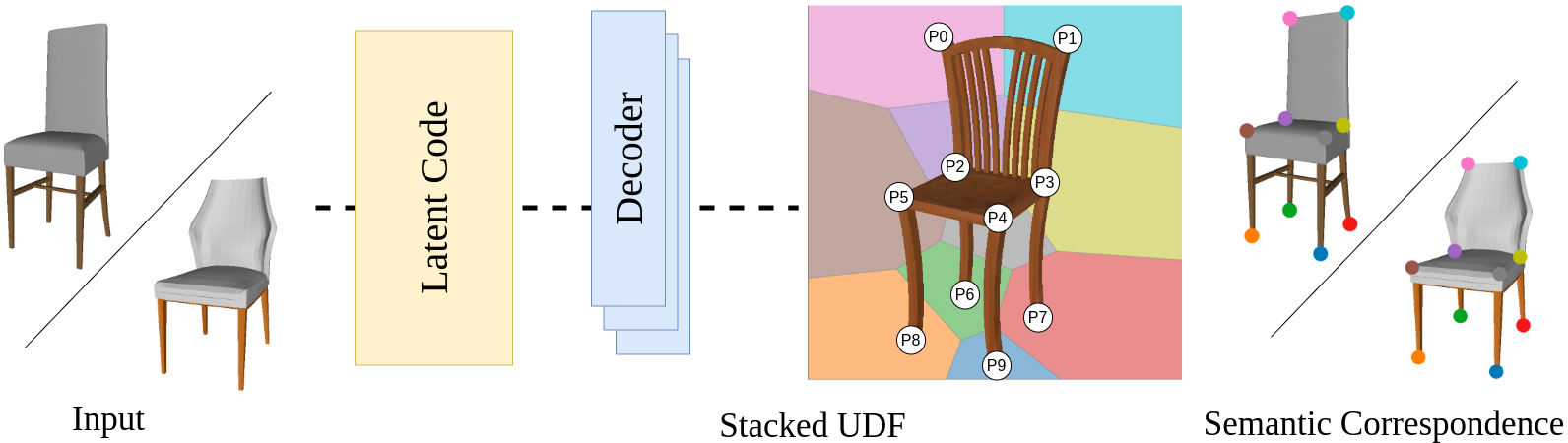}
	\caption{Illustration of the stacked UDF for semantic label prediction. In this case, $\forall \point \in \mathbb{R}^3$, the stacked UDF of $\point$ is $[d_i]_{i=0}^9$, $d_i = \left \| \point - P_i\right \|_2, i=0,...,9$, where $P_i$ is a specific keypoint. Then $\text{Label}(\point) = \arg \min_i [d_i]_{i=0}^9$. We use a 2D Voronoi diagram to illustrate our learned stacked UDF, where the painted region represents points with the same semantic label. }
	\label{fig:udf_demo}
	\vspace{-5mm}
\end{figure}

\subsection{Implicit Semantic Learning}
\label{sec:3.3}

In the previous section, we proposed a method to predict the location of keypoints. Practically, the semantic labels (e.g., the head of an airplane, the foot of a chair) of keypoints also play a very important role, such as building correspondence across diverse shapes in the same category. Taking the predicted keypoints of the above framework as input, the semantic keypoint labeling can be treated as a per-point classification task. 

For each keypoint, a straightforward way is to extract features conditioned on its coordinates and neighborhood and to feed into an MLP-based classifier. Since the input is a complete point cloud, this task is similar to conducting point-wise classification for picking up the semantic keypoints, such as the method RSNet~\cite{huang2018recurrent} used in \cite{you2020keypointnet}. However, it is still very challenging when the input is a partial point cloud or a single-view image, due to the difficulty to provide sufficient features for some unobservable keypoints. 

Encouraged by the success of implicit representation learning for keypoint estimation, we also explore an implicit way to do semantic keypoint learning. Inspired by the stacked fashion of heatmap-based representation~\cite{newell2016stacked}, we propose our stacked UDF representation. Specifically, assuming the number of semantic labels is $K$, instead of placing $K$ keypoint labels at one channel of space, we distribute them to $K$ channels of spaces, with one label per channel. For each channel, the distance values from quired points to the specific keypoint of the channel can be computed, as the UDF values.

\paragraph{Stacked UDF learning.}
We use a neural network to fit stacked UDF for implicit semantic label learning, as shown in Figure~\ref{fig:udf_demo}. Specifically, for an arbitrary point $\point$ in the space, the  continuous stacked UDF $g$ can be defined as:
\begin{align}
    g(\point): \mathbb{R}^3 \rightarrow d^K \in \mathbb{R}^K,
\end{align}
where $K$ is the maximum number of keypoint labels, $d^K$ represents the distance from $\point$ to the corresponding keypoint of every channel. Note that the distance value of the channel should be infinite when the semantic label does not exist. In our experiments, we simply set 1 as the value of the nonexistent label. Given an observation $x\in\mathcal{X}$, we have the conditional network function:
\begin{align}
    g_\theta: \mathbb{R}^3 \times \mathcal{X} \rightarrow d^K \in \mathbb{R}^K.
\end{align}
The stacked UDF network shares the same structure as the SDF branch except for the last linear layer in the MLP decoder. This UDF branch is trained using the supervision of $L_1$ loss on UDF values.

\paragraph{Semantic label prediction.}
Given the keypoints estimated in Section~\ref{sec:3.1}, we can obtain their corresponding ${d^K}$ with learned stacked UDF. Labels of keypoints can then be obtained by using an `$\mathrm{argmin}$' operation to pick out the channel which owns the minimum distance. Notice our method is invariant to the order of input keypoints, making it suitable for unordered keypoint input. 


In our semantic learning, we utilize UDF instead of SDF for several reasons. Although using UDF may have a slightly larger fitting error for keypoints compared to using SDF, it still yields comparable results in the task of semantic learning. This is because the accuracy of label prediction is not highly sensitive to the fitting error. Additionally, UDF offers advantages in terms of implementation simplicity, data preparation, and faster inference speed.

Note that in our approach, we separate the tasks of keypoint estimation and semantic learning to achieve optimal performance. An alternative solution is jointly learning keypoint spheres and semantic labels using a similar stacked SDF representation. In this representation, each channel corresponds to the SDF of keypoints associated with a specific label. However, training a stacked SDF model learning requires much more data sampling to ensure accurate keypoint estimation for all channels, significantly increasing the consumption of calculation and storage. Joint learning also may bring more artifacts thus possibly increasing the difficulty of training and the subsequent keypoint extraction. Therefore, we choose to divide the problem into two subtasks to ensure optimal performance.

\section{Experiment Results}
\label{sec:experiments}
In this section, we introduce the dataset and implementation details in our experiments (Section~\ref{sec:experimental_setup}) and evaluate our implicit keypoint learning scheme on the task of keypoint detection (Section~\ref{sec:exp_keypoint_detection}), keypoint estimation (Section~\ref{sec:exp_keypoint_estimation}), and semantic label inference (Section~\ref{sec:exp_semantic_learning}). An ablation study is also conducted for analyzing our architectural design (Section~\ref{sec:exp_ablation}).

\begin{table*}[htb]
    \centering
    \begin{adjustbox}{width=1.67\columnwidth,center}
        \begin{tabular}{l|c|c|c|c|c|c|c|c|c|c|c}
        \hline
        Method & Metric & Airplane & Bath & Chair & Car & Guitar & Knife & Laptop & Motor & Table & Vessel \\
        \hline
        \multirow{2}{4em}{PointNet} &BHD & 0.366 & 0.422 & 0.310 & 0.474 & 0.612 & 0.890 & 1.022 & 0.542 & 0.660 & 0.360 \\
        &CD & 0.070 & 0.081 & 0.097 & 0.198 & 0.249 & 0.376 & 0.552 & 0.276 & 0.253 & 0.110 \\
        \hline
        \multirow{2}{4em}{DGCNN} &BHD & 0.321 & 0.354 & 0.421 & 0.247 & 0.183 & 0.775 & 0.635 & 0.474 & 0.159 & 0.320 \\ 
        &CD & 0.098 & 0.156 & 0.119 & 0.023 & 0.019 & 0.113 & 0.433 & 0.166 & 0.073 & 0.054 \\
        \hline 
        \multirow{2}{4em}{Ours} &BHD &\textbf{0.124} &\textbf{0.235} &\textbf{0.148} &\textbf{0.121} &\textbf{0.097} &\textbf{0.147} &\textbf{0.097} &\textbf{0.194} &\textbf{0.117} &\textbf{0.260} \\
        &CD &\textbf{0.015} &\textbf{0.031} &\textbf{0.018} &\textbf{0.009} &\textbf{0.007} &\textbf{0.025} &\textbf{0.015} &\textbf{0.017} &\textbf{0.026} &\textbf{0.053} \\
        \hline
        \end{tabular}
    \end{adjustbox}
    \caption{Comparison results of PointNet~\cite{qi2017pointnet}, DGCNN~\cite{wang2019dynamic}, and Ours using complete point cloud input. Average BHD and CD are reported, the lower value is better. Our method utilizes the same encoder as PointNet while adopting different keypoint representations. DGCNN is the best keypoint saliency benchmark in KeypointNet~\cite{you2020keypointnet}.}
    \label{tab:kp_sailency}
    \vspace{-3mm}
\end{table*}
\begin{table*}[htb]
    \centering
    
    \begin{adjustbox}{width=1.67\columnwidth,center}
        \begin{tabular}{ l|c|c|c|c|c|c|c|c|c|c|c }
        \hline
        Method & Metric & Airplane & Bath & Chair & Car & Guitar & Knife & Laptop & Motor & Table & Vessel \\
         \hline
        \multirow{2}{4em}{Coords} &BHD &0.190 & 0.249 & 0.245 & 0.154 & 0.126 & 0.174 & 0.127 & 0.219 & 0.133 & 0.281 \\
        &CD &0.028 & 0.047 & 0.060 & 0.013 & 0.012 & 0.036 & 0.025 & 0.025 & \textbf{0.025} & 0.060 \\
        \hline
        \multirow{2}{4em}{Heatmap} &BHD &0.241 & 0.474 & 0.366 & 0.163 & 0.189 & 0.248 & 0.834 & 0.256 & 0.213 & 0.611 \\
        &CD &0.086 & 0.345 & 0.196 & 0.018 & 0.030 & 0.076 & 1.315 & 0.049 & 0.148 & 0.600 \\
        \hline
        \multirow{2}{4em}{Ours} &BHD &\textbf{0.124} & \textbf{0.235} & \textbf{0.148} & \textbf{0.121} & \textbf{0.097} & \textbf{0.147} & \textbf{0.097} & \textbf{0.194} & \textbf{0.117} & \textbf{0.260} \\
        &CD &\textbf{0.015} & \textbf{0.031} & \textbf{0.018} & \textbf{0.009} & \textbf{0.007} & \textbf{0.025} & \textbf{0.015} & \textbf{0.017} & 0.026 & \textbf{0.053} \\
        \hline
        \end{tabular}
    \end{adjustbox}
    \caption{Comparison results of coordinate regression, heatmap inference, and Ours using complete point cloud input. Average BHD and CD are reported, the lower value is better. }
    \label{tab:full_pc}
    \vspace{-5mm}
\end{table*}

\subsection{Experimental Setup}
\label{sec:experimental_setup}

\paragraph{Dataset.}
We use the KeypointNet dataset~\cite{you2020keypointnet} which contains 103,450 annotated keypoints and 8,234 3D models spanning 16 object categories from ShapeNet~\cite{chang2015shapenet}. We choose 10 popular categories, i.e., the airplane, bathtub, car, chair, guitar, knife, laptop, motorcycle, table, and vessel, for our experiments with point cloud input, and pick rendered images of the corresponding categories in the dataset of 3D-R2N2 \cite{choy20163d} for the experiments using image input. We randomly split the data into a train set (80\%) and a test set (20\%). More specifically, all the 3D models are normalized into a bounding box of $[-1,1]^3$. To acquire data pairs for training SDF fields of keypoints, we first represent keypoints as spheres of radius $0.08$, created in MeshLab with 2,562 vertices. Subsequently, we uniformly sample 100,000 points in $[-1,1]^3$ and collect the $N_S \times 2,562$ ($N_S$ is the keypoint number of $S$) sphere surface points, along with their corresponding SDF values, respectively. For data in semantic learning, samples are kept the same with the SDF setting, and the corresponding distances (i.e., UDF values) to all keypoints will be calculated for each sample point.

\paragraph{Network and training.}
For the network structure,  we deploy classical encoders w.r.t diverse kinds of input followed by the same decoder to estimate keypoints. Specifically, we employ ResNet~\cite{he2016deep} for single image input and PointNet~\cite{qi2017pointnet} for point cloud input. The dimension of the last layer in all encoders is set to 256. We also incorporate the positional encoding, generating a $39d$ (where $d$ means dimension) feature for each queried point in space. This feature is then concatenated with the $256d$ feature encoded from the input. Subsequently, the $295d$ feature is forwarded to the implicit decoder. The decoder is an MLP-based network consisting of 5 fully connected layers with a sine activation between layers. The output channels are 256, 256, 256, 256, and 1, respectively. During the training of the SDF field, we randomly generate 10,000 volume-based samples and 10,000 surface-based ones (if insufficient, we just add those uniform samples) for each shape. The batch size is set to 4. For training details, in our experiments, we set $\omega=30$ in the sine function and use $\lambda_1=1.0, \lambda_2=0.1, \lambda_3 = 0.05$ in the weighted training loss function $L$. Adam optimizer is adopted to train our network with an initial learning rate of 1e-4, $\beta_1=0.9$, and $\beta_2=0.999$. We train our network with 300 epochs for all tasks on one GTX 2080ti GPU.

\paragraph{Evaluation metrics.}
The Chamfer Distance (CD) for traditional point cloud generation work \cite{fan2017point} is adopted to evaluate the distance between the predicted keypoints and the ground truth where the number of points might be different. We also utilize Bidirectional Hausdorff Distance (BHD) to measure the similarity between two point sets. Suppose $S_1, S_2$ are two point sets, BHD ($d_H$) and CD ($d_C$) are defined as:
\begin{align}   \label{BHD}
    d_{H}(S_1, S_2) &= \frac{1}{2}(\max_{p \in S_1} \min_{q \in S_2} \left \|p - q \right \|_2 \\ \nonumber
    & + \max_{q \in S_2} \min_{p \in S_1} \left \|p - q \right \|_2 ) \\
    d_{C}(S_1, S_2) &= \frac{1}{|S_1|} \sum_{p \in S_1} \min_{q \in S_2} \left \|p - q \right \|_2^2 \\ \nonumber
    &+ \frac{1}{|S_2|} \sum_{q \in S_2} \min_{p \in S_1} \left \|p - q \right \|_2^2.
\end{align}

\subsection{Comparisons on Keypoint Detection}
\label{sec:exp_keypoint_detection}

Although our method focuses on keypoint estimation of general objects, especially for incomplete input, we first compare our method with state-of-the-art methods in the field of keypoint detection. Table~\ref{tab:kp_sailency} presents the results of our method, PointNet~\cite{qi2017pointnet} (using the same encoder as ours), and DGCNN~\cite{wang2019dynamic} (the best of keypoint saliency benchmark in KeypointNet). Both of these classification methods have a tendency to predict an excessive number of keypoints, resulting in significant errors in BHD and CD metrics. Our method outperforms them significantly in terms of BHD and CD, showcasing its superiority.

\begin{figure}
	\centering
	\includegraphics[width=\linewidth]{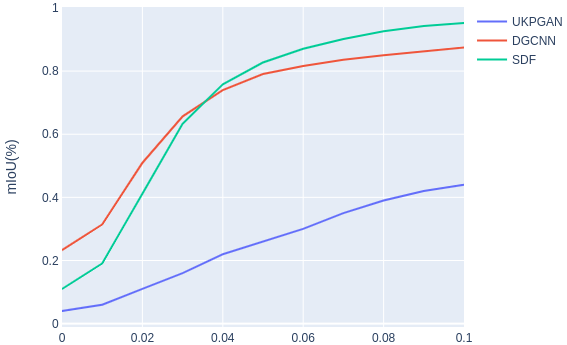}
	\caption{The mIoU results under various distance thresholds (0-0.1) for compared algorithms, i.e., our method (SDF), DGCNN~\cite{wang2019dynamic}, and UKPGAN~\cite{you2020ukpgan}. Note that our method is significantly better than UKPGAN and outperforms DGCNN when the threshold is larger than 0.04.}
	\label{fig:mIoU}
	\vspace{-5mm}
\end{figure}

We further report mIoU curves in line with KeypointNet~\cite{you2020keypointnet} in Figure~\ref{fig:mIoU}. Our method outperforms DGCNN when the distance threshold is bigger than 0.04. It also achieves better results than an unsupervised keypoint generation method, i.e. UKPGAN~\cite{you2020ukpgan}. The lower mIoU of our method at a small threshold is basically caused by the error of off-surface distance. 

\begin{figure*}[t]
	\centering
	\includegraphics[width=0.95\linewidth]{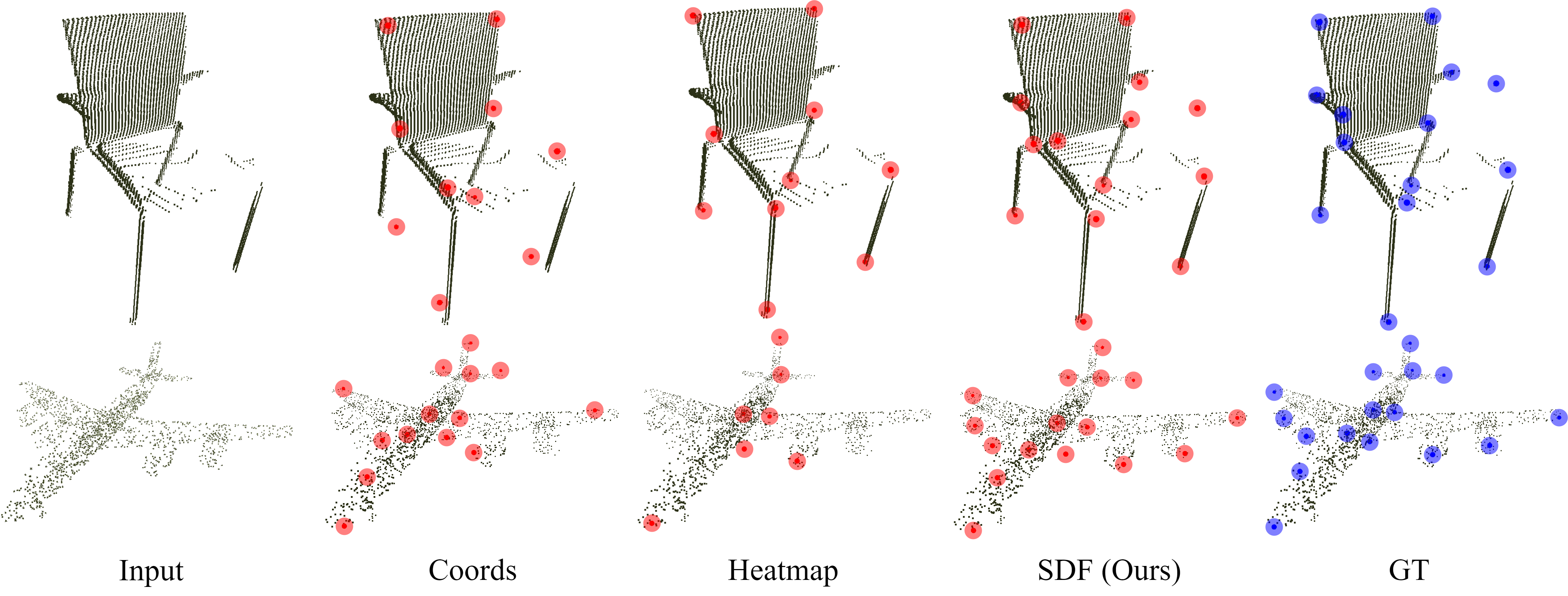}
	\caption{Qualitative results of keypoint estimation with a point cloud input. The input of the first and second rows are a partial point cloud and a complete point cloud, respectively. `Coords' means coordinate regression, and `Heatmap' means heatmap inference.}
	\label{fig:pcd_comp}
	\vspace{-2mm}
\end{figure*}

\begin{figure*}[t]
	\centering
	\includegraphics[width=0.9\linewidth]{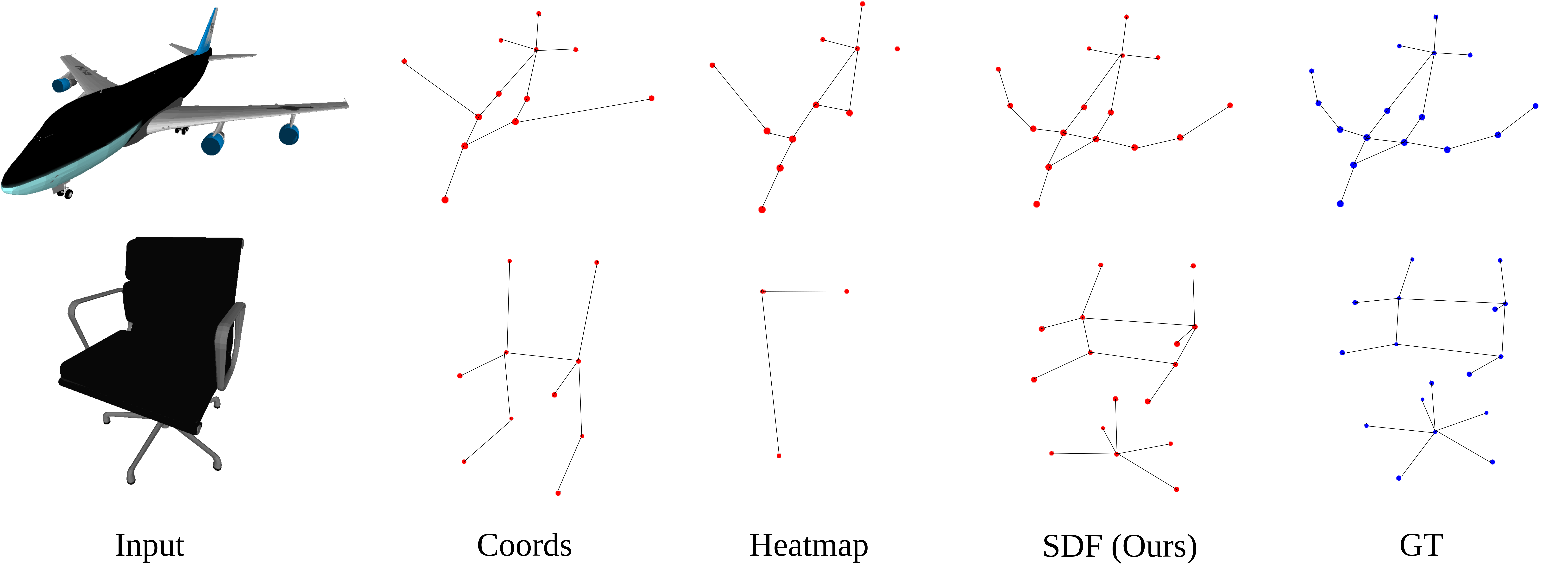}
        \vspace{-3mm}
	\caption{Qualitative results of keypoint estimation with a single-view image input. `Coords' means coordinate regression, and `Heatmap' means heatmap inference. For better visualization, we draw lines among the predicted keypoints.}
	\label{fig:img_comp}
\end{figure*}

\begin{table*}[htb]
    \centering
    \begin{adjustbox}{width=1.68\columnwidth,center}
    \begin{tabular}{ l|c|c|c|c|c|c|c|c|c|c|c }
        \hline
        Method & Metric & Airplane & Bath & Chair & Car & Guitar & Knife & Laptop & Motor & Table & Vessel \\
         \hline
        \multirow{2}{4em}{Coords} &BHD &0.171 & \textbf{0.209} & 0.196 & 0.135 & 0.135 & \textbf{0.144} & \textbf{0.084} & 0.201 & \textbf{0.105} & \textbf{0.294} \\
        &CD &0.025 & \textbf{0.036} & 0.036 & 0.010 & 0.013 & \textbf{0.023} & \textbf{0.010} & 0.020 & \textbf{0.019} & \textbf{0.068} \\
        \hline
        \multirow{2}{4em}{Heatmap} &BHD &0.236 & 0.359 & 0.412 & 0.141 & 0.155 & 0.388 & 0.616 & 0.315 & 0.433 & 0.504 \\
        &CD &0.097 & 0.248 & 0.279 & 0.01 & 0.026 & 0.206 & 0.79 & 0.087 & 0.456 & 0.43 \\
        \hline
        \multirow{2}{4em}{Ours} &BHD &\textbf{0.136} & \textbf{0.209} & \textbf{0.186} & \textbf{0.111} & \textbf{0.097} & 0.149 & 0.111 & \textbf{0.187} & 0.124 & 0.331 \\
        &CD &\textbf{0.020} & \textbf{0.036} & \textbf{0.032} & \textbf{0.007} & \textbf{0.007} & 0.030 & 0.051 & \textbf{0.017} & 0.029 & 0.123 \\
    
        \hline
    \end{tabular}
    \end{adjustbox}
    \caption{Comparison results of coordinate regression, heatmap inference, and Ours using partial point cloud input. Average BHD and CD are reported, the lower value is better.}
    \label{tab:partial_pc}
    \vspace{-5mm}
\end{table*}

\subsection{Comparisons on Keypoint Estimation}
\label{sec:exp_keypoint_estimation}

To validate the effectiveness of our proposed implicit learning approach for 3D keypoint estimation, we conduct experiments on general objects using different input types, i.e., complete point clouds, partial point clouds, and single-view images. The baselines are based on coordinate regression and heatmap inference methods. Because the output dimension of the point regression network is forced to be fixed, we adapt it by learning a binary mask for each predicted keypoint, making it usable for number-varying point prediction. For the heatmap-based method, we represent all keypoints in a 3D heatmap ($128^3$) to alleviate the calculation and storage consummation. To ensure fairness and efficiency, we employ the same encoder for different methods. Specifically, PointNet~\cite{qi2017pointnet} and ResNet18~\cite{he2016deep} are chosen for point cloud and image input, respectively.

\paragraph{Results on complete point clouds.}
For the complete point cloud input, we employ the same encoder, i.e. PointNet, for different methods. Our method performs the best across almost all categories for all metrics, as demonstrated in Table~\ref{tab:full_pc}. Visual results are also shown in Figure~\ref{fig:pcd_comp} (the second row).

\paragraph{Results on partial point clouds.}
We render 24 views of depth maps for each mesh in the KeypointNet dataset~\cite{you2020keypointnet} and then obtain partial point clouds (following the approach used in ME-PCN~\cite{gong2021me}) to evaluate the robustness of different methods. The network structures remain unchanged compared to the experiments of complete point clouds. Our method also exhibits its superiority in most categories, seen in Table~\ref{tab:partial_pc} and Figure~\ref{fig:pcd_comp} (the first row). However, when dealing with partial data, both the heatmap method and our approach encounter challenges in accurately predicting the number of keypoints, which can result in potentially larger errors.

\begin{figure*}[t]
	\centering
	\includegraphics[width=\linewidth]{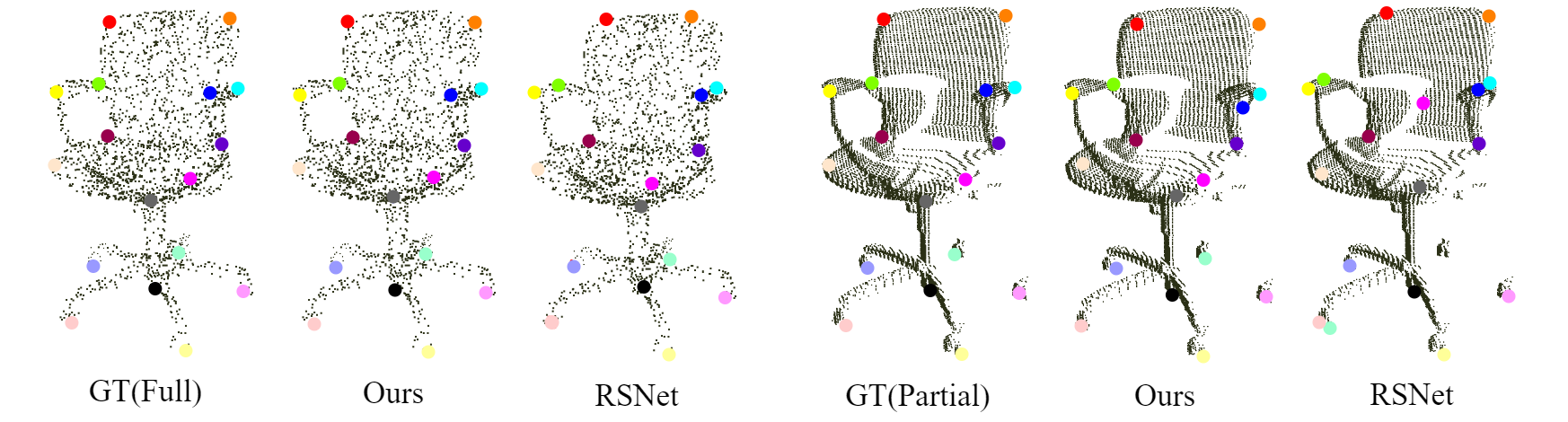}
	\caption{Visualization for results of semantic learning. Left: semantic label prediction with full point cloud input; Right: prediction with the partial point cloud. Our method predicts semantic labels of estimated keypoints by the proposed stacked UDF, while RSNet~\cite{huang2018recurrent} performs classification on the input point cloud. As seen on the right side, RSNet may generate close keypoints with different semantic labels and keypoints in incorrect positions due to missing data in the input.}
	\label{fig:semantic_vis}
\end{figure*}

\begin{table}[t]
    \centering
        \begin{tabular}{ l|c|c|c|c|c|c }
        \hline
        Method & Metric & Airplane & Chair & Car & Table & Vessel \\
         \hline
        \multirow{2}{4em}{Coords} &BHD &0.247 & \textbf{0.145} & 0.259 & \textbf{0.140} & 0.334 \\
        &CD &0.052 & 0.071 & 0.013 & 0.035 & 0.092 \\
        \hline
        \multirow{2}{4em}{Heatmap} &BHD &0.389 & 0.602 & 0.215 & 0.652 & 0.734 \\
        &CD &0.181 & 0.564 & 0.045 & 1.079 & 0.545 \\
        \hline
        \multirow{2}{4em}{Ours} &BHD &\textbf{0.217} & 0.214 & \textbf{0.136} & \textbf{0.140} & \textbf{0.310} \\
        &CD &\textbf{0.038} & \textbf{0.049} & \textbf{0.012} & \textbf{0.032} & \textbf{0.078} \\
    
        \hline
        \end{tabular}
    \caption{Comparison results of coordinate regression, heatmap inference, and Ours using single-view image input. Average BHD and CD are reported, the lower value is better.}
    \label{tab:sv_img}
\end{table}

    


\paragraph{Results on single-view images.}
Similar to single-view reconstruction, the goal of this task is to accurately estimate the complete set of keypoints for a given single image. We evaluate different methods on 5 object categories, in which the rendering images come from 3D-R2N2~ \cite{choy20163d}. Quantitative and qualitative results are shown in Table~\ref{tab:sv_img} and Figure~\ref{fig:img_comp}. Our method outperforms others, and the discussion of the partial data effect is similar to the case of the partial point cloud.

\subsection{Comparisons on Semantic Learning}
\label{sec:exp_semantic_learning}

In this paper, we introduce an exploring method of semantic label learning for the keypoints generated by our algorithm. Figure~\ref{fig:semantic_vis} illustrates a visual comparison with RSNet~\cite{huang2018recurrent}, which is the best keypoint correspondence benchmark in KeypointNet~\cite{you2020keypointnet}. We evaluate our method using two types of inputs, i.e. full and partial point clouds. To enhance visualization, keypoints with distinct semantic labels are colored in different ways. Our method achieves accurate semantic correspondence with ground truth (GT) for both full and partial point cloud inputs, although there might be some keypoint errors in the case of partial inputs. RSNet demonstrates comparable performance on the full point clouds, but it can have two keypoints with different labels close to each other in the case of partial point cloud inputs. Our method exhibits superiority in handling partial data inputs.

\begin{table}
    \centering
        \begin{tabular}{l|c|c|c}
        \hline
         Metric & Num & BHD & CD \\
        \hline
        SDF & \textbf{8} & \textbf{0.0024} & \textbf{1.3e-5} \\
        \hline
        UDF &7 & 0.0692 & 0.0119 \\
        \hline
        \end{tabular}
    \caption{Comparisons of SDF and UDF learned on an example. The ground-truth number of keypoint is 8. The lower of BHD or CD, the result is better.}
    \label{tab:sdf_udf}
\end{table}

\begin{table}[ht]
    \centering
        \begin{tabular}{l|c|c|c|c|c}
        \hline
          Radius & 0.24 & 0.16 & 0.08 & 0.04 & 0.02\\
        \hline
        BHD & 0.1507 &\textbf{0.1470} & 0.1472 & 0.2148 & 0.3029 \\
        CD & 0.0192 & 0.0181 & \textbf{0.0173} & 0.0411 & 0.0696 \\
        \hline
        \end{tabular}
    \caption{Ablation study of sphere radius. The lower of BHD or CD, the result is better.}
    \label{tab:radius_sphere}
\end{table}

\begin{table*}[htb]
    \centering
    
    \begin{adjustbox}{width=1.8\columnwidth,center}
        \begin{tabular}{ l|c|c|c|c|c|c|c|c|c|c|c|c }
        \hline
        Method & Metric & Airplane & Bath & Chair & Car & Guitar & Knife & Laptop & Motor & Table & Vessel & Mean \\
         \hline
        \multirow{3}{4em}{RSNet} & Top-1 & 49.54 & 65.14 & 89.40 & 60.40 & 89.78 & 51.25 & 96.21 & 62.77 & 95.47 & 80.35 & 74.03 \\ 
        & Top-3 & 71.99 & 87.06 & 97.97 & 88.61 & 98.63 & 90.00 & 100.00 & 81.55 & 99.39 & 96.09 & 91.13 \\ 
        & Top-5 & 77.90 & 92.62 & 99.30 & 95.49 & 99.36 & 100.00 & 100.00 & 95.05 & 99.73 & 98.41 & \textbf{95.79} \\
        \hline
        \multirow{3}{4em}{Ours} &Top-1 & 76.78 & 58.17 & 93.44 & 80.87 & 86.09 & 67.50 & 100.00 & 57.16 & 99.18 & 49.07 & \textbf{76.82}\\
        & Top-3 & 92.47 & 82.32 & 98.40 & 96.10 & 98.06 & 91.67 & 100.00 & 79.66 & 99.62 & 77.52 & \textbf{91.58}\\
        & Top-5 & 96.02 & 86.43 & 98.94 & 98.74 & 98.72 & 98.75 & 100.00 & 87.47 & 99.66 & 83.23 & 94.80\\
        \hline
        \end{tabular}
    \end{adjustbox}
    \caption{Comparison results of stacked UDF (Ours) and RSNet~\cite{huang2018recurrent} with complete point cloud input. The Top-1, Top-3, and Top-5 accuracy are reported. Our method performs slightly better than RSNet. }
    \label{tab:semantic_full_pc}
    \vspace{-3mm}
\end{table*}
\begin{table*}[htb]
    \centering
        \begin{tabular}{ l|c|c|c|c|c|c|c|c|c|c|c|c }
        \hline
        Method & Metric & Airplane & Bath & Chair & Car & Guitar & Knife & Laptop & Motor & Table & Vessel & Mean \\
        \hline
        \multirow{3}{4em}{RSNet} & Top-1 & 34.19 & 40.60 & 58.66 & 49.34 & 69.53 & 54.79 & 80.49 & 48.19 & 77.96 & 63.39 & 57.71 \\
        &Top-3 & 58.37 & 64.82 & 78.19 & 78.28 & 89.05 & 88.13 & 99.43 & 72.63 & 97.17 & 87.84 & 81.39 \\
        &Top-5 & 67.26 & 76.44 & 86.29 & 88.58 & 94.29 & 99.16 & 100.00 & 84.47 & 98.90 & 94.59 & 89.00 \\
        \hline
        \multirow{3}{4em}{Ours} &Top-1 & 62.65 & 49.57 & 91.86 & 74.75 & 84.45 & 64.38 & 100.00 & 56.86 & 98.07 & 43.71 &\textbf{72.63}\\
        &Top-3 & 83.29 & 78.54 & 97.59 & 87.87 & 96.82 & 91.67 & 100.00 & 75.88 & 99.57 & 74.18 & \textbf{88.54}\\
        &Top-5 & 88.90 & 82.50 & 98.45 & 93.58 & 98.19 & 99.17 & 100.00 & 82.45 & 99.66 & 82.61 & \textbf{92.55}\\
        \hline
        \end{tabular}
    \caption{Comparison results of stacked UDF (Ours) and RSNet~\cite{huang2018recurrent} with partial point cloud input. The Top-1, Top-3, and Top-5 accuracy are reported. It shows that our method is much more robust than RSNet for partial point cloud input. }
    \label{tab:semantic_partial_pc}
\end{table*}

We also present a quantitative comparison between our stacked UDF and RSNet~\cite{huang2018recurrent} on the settings of complete point cloud input (Table~\ref{tab:semantic_full_pc}) and partial point cloud input (Table~\ref{tab:semantic_partial_pc}). As shown in these two tables, the Top-1, Top-3, and Top-5 accuracy are reported. RSNet~\cite{huang2018recurrent} can achieve comparable performance with ours for complete point cloud input but it degenerates dramatically for partial point cloud input. In contrast, our method can robustly predict semantic labels for both complete and partial data, as stated in Section~\ref{sec:3.3} of our paper.

\begin{table*}[ht]
    \centering
        \begin{tabular}{l|c|c|c|c|c|c|c|c|c|c|c|c}
        \hline 
        Activation & \multicolumn{4}{|c|}{ReLU} & \multicolumn{4}{|c|}{SeLU} & \multicolumn{4}{|c}{Sine} \\
        \hline
            Pos, Grad & wo,wo & wo,w/ & w/,wo & w/,w/ & wo,wo & wo,w/ & w/,wo & w/,w/ & wo,wo & wo,w/ & w/,wo & w/,w/ \\
        \hline
            BHD &0.284 & 0.308 & 0.227 & 0.188 & 0.237 & 0.207 & 0.243 & 0.171 & 0.216 & 0.806 & 0.204 & \textbf{0.148} \\
            CD &0.050 & 0.057 & 0.038 & 0.028 & 0.043 & 0.036 & 0.049 & 0.024 & 0.031 & 0.321 & 0.027 & \textbf{0.018} \\ 
        \hline
        \end{tabular}
    \caption{Ablation study of network architecture. We evaluate the effectiveness with (w/) or without (wo) positional encoding (Pos) and gradient loss (Grad), as well as different activation functions (ReLU, SeLU, and Sine). The lower of BHD or CD, the result is better.}
    \label{tab:net_params}
\end{table*}

\begin{figure}
    \centering
    \includegraphics[width=1.\linewidth]{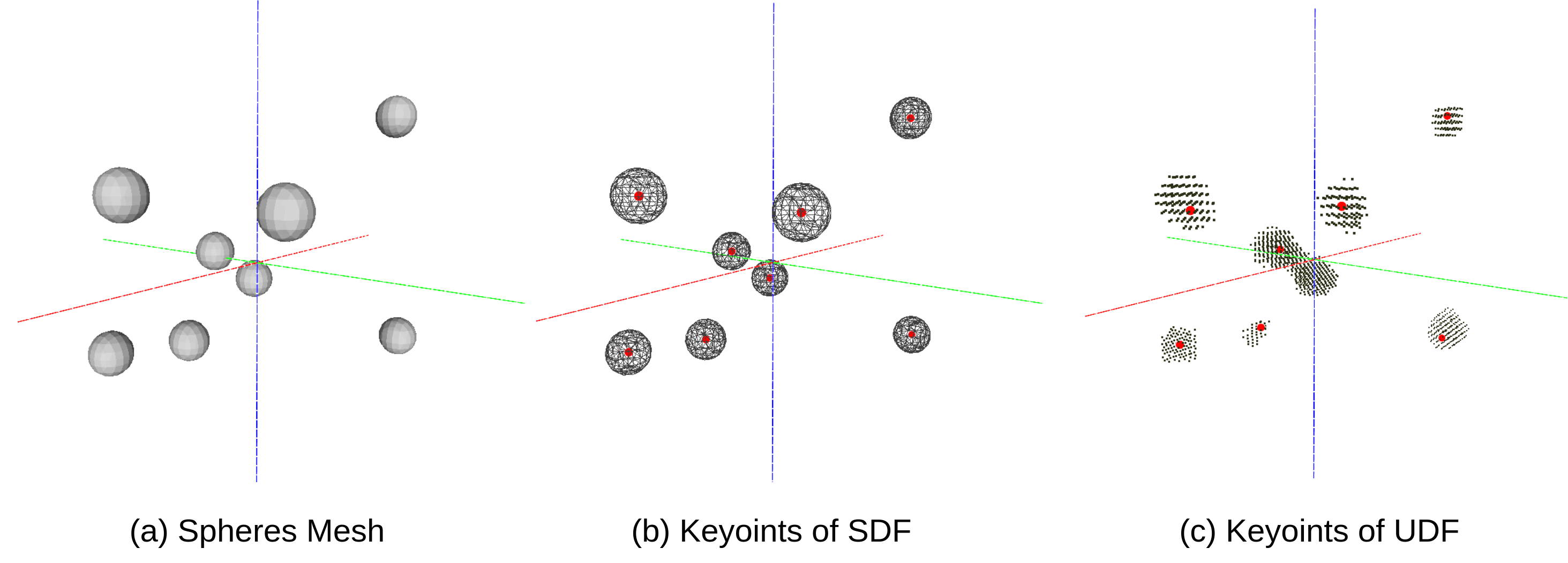}
    \caption{Visualization of fitting results of SDF and UDF. (a) is the spheres' mesh of SDF by Marching Cubes~\cite{lorensen1987marching}. (b) shows the extracted keypoints from the spheres' mesh. (c) shows points whose UDF values are larger than the threshold 0.08 and extracts final keypoints by an `argmin' function. } 
    \label{fig:sdf_vs_udf}
    \vspace{-5mm}
\end{figure}

\begin{figure}[t]
	\centering
	\includegraphics[width=\linewidth]{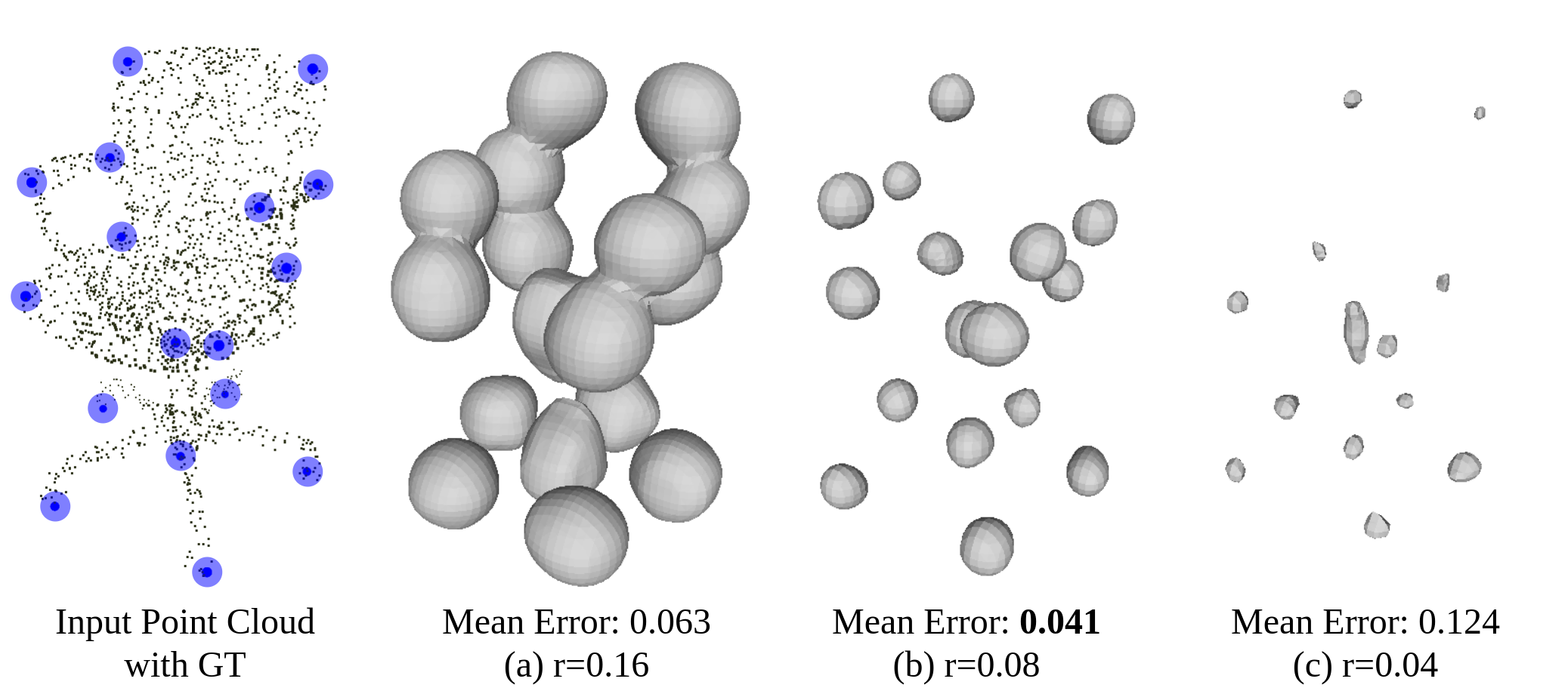}
	\vspace{-3mm}
	\caption{Ablation study about the choice of sphere radius. In this case, mean $L_2$ distances of keypoints are reported. As shown in the figure, adopting a small radius tends to generate bad shapes. Meanwhile, a large radius leads to generating intersected spheres, increasing the difficulty of keypoint extraction.}
	\label{fig:radis_abl}
	\vspace{-5mm}
\end{figure}

\subsection{Ablation Study}
\label{sec:exp_ablation}

In this subsection, we provide an ablation study to validate the effectiveness of our architectural design and hyper-parameter selection. For simplicity without loss of generality, the study is performed on the chair category with complete point cloud input.

\paragraph{Network architecture.}
We investigate the impact of using positional encoding, gradient loss, and different activation functions in our network training. The results are provided in Table~\ref{tab:net_params}, showing the network with positional encoding, gradient loss, and sine activation performs the best. Positional encoding helps the network learn high-frequency features in the data as stated in NeRF~\cite{mildenhall2020nerf} and sine activation combined with gradient loss can improve the smoothness of underlying surface discussed in SIREN~\cite{sitzmann2020implicit}. Consequently, these architectural choices contribute to improved keypoint extraction and yield superior results.

\paragraph{SDF vs UDF.}
As stated in Section~\ref{sec:3.1}, we adopt SDF for keypoint learning. To compare the regression capabilities of SDF and UDF representations, we use the same MLP network to fit the SDF and UDF fields of 8 points randomly sampled from the $[-1,1]^3$ space. The numerical results are presented in Table~\ref{tab:sdf_udf}. We also give a visualization for the qualitative comparison in Figure~\ref{fig:sdf_vs_udf}. For UDF learning, we extract the underlying keypoints from the clusters if their values are larger than the given threshold of 0.08. However, the cluster shapes are not as good as the output of our SDF learning, making the UDF keypoints extracted from `argmin' function deviate from the ground truth positions.

\paragraph{Sphere radius.}
The performance of our SDF representation is influenced by the choice of sphere radius. As shown in Table~\ref{tab:radius_sphere}, using a small radius data or a large radius degrades the performance of keypoint estimation. With a small radius, the spheres cannot be well-fitted by the network probably due to local imbalances of SDF values and some numerical issues. On the other hand, using a large radius increases the likelihood of sphere intersections, which complicates keypoint extraction and slows down the process. Figure~\ref{fig:radis_abl} provides visual results to support these observations. Therefore, we adopt a default sphere radius of 0.08, which remains fixed throughout all our experiments.



\section{Conclusion}
\label{sec:conclusion}

In this paper, we propose a novel framework for general object keypoint estimation, which is the first attempt to introduce continuous implicit field learning into the prediction of sparse and distinct points. It addresses the challenges related to the uncertain number and order properties of keypoints and enhances the performance of 3D keypoint estimation on incomplete input, including partial point clouds and single-view images. Moreover, the proposed implicit representation facilitates semantic label inference. Experimental results demonstrate that our novel keypoint estimation formulation surpasses existing methods that rely on position regression and heatmap inference techniques.


In terms of limitations and future work, our method utilizes a predefined sphere radius for implicit field calculation. It would be valuable to explore the potential benefits of adaptively adjusting the radius based on the specific object categories, which could potentially enhance the accuracy and robustness of keypoint estimation. Additionally, we are intrigued by the prospect of studying dense point scenarios that pose greater challenges for network learning and keypoint extraction. Moreover, an interesting avenue for future research is to develop an end-to-end architecture that directly obtains keypoints from an SDF-based representation, eliminating the need for an intermediate step. Lastly, our method is a supervised approach that relies on an annotated keypoint dataset. In the future, we would like to study unsupervised learning for keypoint estimation which can be generalized to unseen object categories.



\vspace{1mm}
\noindent\textbf{Acknowledgment.} 
This work was partially supported by NSFC-62172348, the Basic Research Project No. HZQB-KCZYZ-2021067 of Hetao Shenzhen-HK S\&T Cooperation Zone, the National Key R\&D Program of China with grant No. 2018YFB1800800, by Shenzhen Outstanding Talents Training Fund 202002, by Guangdong Research Projects No. 2017ZT07X152 and No. 2019CX01X104, by the Guangdong Provincial Key Laboratory of Future Networks of Intelligence (Grant No. 2022B1212010001), and by Shenzhen Key Laboratory of Big Data and Artificial Intelligence (Grant No. ZDSYS201707251409055). It was also supported in part by Outstanding Yound Fund of Guangdong Province with No.  2023B1515020055 and Shenzhen General Project with No. JCYJ20220530143604010. It was also sponsored by CCF-Tencent Open Research Fund.

\bibliographystyle{eg-alpha-doi}

\bibliography{reference}

\newpage

\end{document}